\documentclass{sig-alternate-05-2015}
\usepackage{enumerate}
\usepackage{subcaption}

\begin{document}

% Copyright
%\setcopyright{acmcopyright}
%\setcopyright{acmlicensed}
%\setcopyright{rightsretained}
%\setcopyright{usgov}
%\setcopyright{usgovmixed}
%\setcopyright{cagov}
%\setcopyright{cagovmixed}

% DOI
\doi{10.1145/2935620.2935622}

% ISBN
\isbn{978-1-4503-4405-0/16/06}

%Conference
\conferenceinfo{DroNet16}{June 26, 2016, Singapore, Singapore}

%\acmPrice{\$15.00}

%
% --- Author Metadata here ---
%\conferenceinfo{WOODSTOCK}{'97 El Paso, Texas USA}
%\CopyrightYear{2007} % Allows default copyright year (20XX) to be over-ridden - IF NEED BE.
%\crdata{0-12345-67-8/90/01}  % Allows default copyright data (0-89791-88-6/97/05) to be over-ridden - IF NEED BE.
% --- End of Author Metadata ---

\title{Generic Drone Control Platform for Autonomous Capture of Cinema Scenes}
\subtitle{}

\numberofauthors{2} %  in this sample file, there are a *total*
\author{
% 1st. author
\alignauthor
Julien Fleureau\\
       \affaddr{Technicolor R\&D Rennes}\\
       \affaddr{975 Avenue des Champs Blancs}\\
       \affaddr{35576 Cesson-S\'evign\'e, France}\\
       \email{julien.fleureau@technicolor.com}
% 2nd. author
\alignauthor
Quentin Galvane\\
       \affaddr{Technicolor R\&D Rennes}\\
       \affaddr{975 Avenue des Champs Blancs}\\
       \affaddr{35576 Cesson-S\'evign\'e, France}\\
       \email{quentin.galvane@technicolor.com}
\and  % use '\and' if you need 'another row' of author names
% 4th. author
\alignauthor
Francois-Louis Tariolle\\
       \affaddr{Technicolor R\&D Rennes}\\
       \affaddr{975 Avenue des Champs Blancs}\\
       \affaddr{35576 Cesson-S\'evign\'e, France}\\
       \email{Francois-Louis.Tariolle@technicolor.com}
% 5th. author
\alignauthor
Philippe Guillotel\\
       \affaddr{Technicolor R\&D Rennes}\\
       \affaddr{975 Avenue des Champs Blancs}\\
       \affaddr{35576 Cesson-S\'evign\'e, France}\\
       \email{philippe.guillotel@technicolor.com}
}
% There's nothing stopping you putting the seventh, eighth, etc.
% author on the opening page (as the 'third row') but we ask,
% for aesthetic reasons that you place these 'additional authors'
% in the \additional authors block, viz.
\additionalauthors{}
\date{19 February 2015}
% Just remember to make sure that the TOTAL number of authors
% is the number that will appear on the first page PLUS the
% number that will appear in the \additionalauthors section.

\maketitle
\begin{abstract}
The movie industry has been using Unmanned Aerial Vehicles as a new tool to produce more and more complex and aesthetic camera shots.
However, the shooting process currently rely on manual control of the drones which makes it difficult and sometimes inconvenient to work with.
In this paper we address the lack of autonomous system to operate generic rotary-wing drones for shooting purposes.
We propose a global control architecture based on a high-level generic API used by many UAV. Our solution integrates a compound and coupled model of a generic rotary-wing drone and a Full State Feedback strategy. % which gives a first-order temporal relation between its flight control and its dynamics. 
To address the specific task of capturing cinema scenes, we combine the control architecture with an automatic camera path planning approach that encompasses cinematographic techniques. 
The possibilities offered by our system are demonstrated through a series of experiments.
%We propose a compound and coupled model of a generic rotary-wing drone which gives a first-order temporal relation between its flight control and its dynamics. 
%describe a global control architecture integrating the two previous models and a Full State Feedback strategy to control the generic rotary-wing drone. 
\end{abstract}



%
%  Use this command to print the description
%

% We no longer use \terms command


\section{Introduction}
With the development of robotics and aeronautics, drones -- also known as Unmanned Aerial Vehicles (UAV) -- are more and more integrated and easy to control, especially in non-military context.
The civil market related to this field is currently experiencing a continuous growth that highlights its potential.
Many applications now make use of such devices for various purposes such as surveillance, communication, video recording or even gaming. 
One particular domain of application has triggered our interest:
with their tradition of using state-of-the-art technology for their production, the cinema industry recently started to exploit the many possibilities offered by these modern flying vehicles.
However, due to the complexity of maneuvering such UAV, one trained human is often still required to pilot the drone, as well as another person to control the embedded camera in case of simultaneous video-recording. Moreover, controlling the drone in terms of position, speed and acceleration as well as in terms of orientation in a very precise way may not be possible, even for an experienced pilot. Such level of control may be especially required when a drone is navigating close to moving persons, or in interior conditions with constrained environments full of 3D obstacles.
Having a solution to autonomously pilot the drone so that it may follow a predefined (or computed in real-time) trajectory is thus definitively useful in such situations.

The challenge that we tackle here is the problem of automatic navigation in a cinematographic context for a regular rotary-wing drone -- the most common type of drone on the civil market. 
Such drones 
often lack a low level API that would allow the direct control of the motors and instead
usually offer a generic way to manually control their trajectory by adapting four parameters: the pitch angle $\theta$ (to move forward and backward), the roll angle $\phi$ (to go left and right), the yaw speed $\dot{\psi}$ (to turn around the vertical axis) and the elevation speed $\dot{z}$ (to move up and down).
A human being has thus to adapt those four parameters in a continuous way to make the drone follow a 6D trajectory (3D for the translations and 3D for the orientation). One can easily understand how hard this task may be in order to produce cinematographic camera trajectories (various speeds, orientations or curves). 
The challenge of making a drone autonomously follow a controlled trajectory falls under the resolution of a control theory problem. Such problem can be classically formalized through the following steps:

\begin{enumerate}
\item Define a dynamic 6D trajectory to be followed by the drone in terms of translation and rotations,
\item Design a model of the drone \emph{i.e.} a mathematic formulation linking the drone inputs (flight control) and its dynamics (position, speed, orientation, ...),
\item Find a way to measure some of the flight dynamics (position, speed, orientation, ...),
\item Build a control system that uses the drone model and the measures of the dynamics to adapt the current flight control so that it follows the given trajectory.
\end{enumerate}

Assuming that the measurement of the flight dynamics is taken care of by a third-party software, we focus on the three remaining steps.
After reviewing the related work on autonomous navigation system for drones and camera path planning techniques, we address the challenges of the second and fourth steps (see section~\ref{servoControl}). We here provide a solution that could be used with any rotary-wing drone providing generic flight control. 
In section~\ref{cinematography} we focus on the first step through the computation of cinematographic trajectories from high-level user inputs.
Finally, we present our results in section~\ref{results} and our leads for future work in section~\ref{limitations}.

\section{Related work}
The research detailed in this paper is multidisciplinary ; it addresses challenges in robotic, path planning and cinematography which all have their own background. In this section, we first review the literature related to autonomous control for drones and then give an overview of the research already conducted in path planning for cinematographic purposes.

\subsection{Autonomous UAV control system}
Drones can be classified under three categories: rotary-wing, fixed-wing and flapping-wing UAV. Due to their stability, maneuverability and hovering capacities, the cinema industry exclusively makes use of rotary-wing drones. Therefore, we decided to limit our research to this specific type of UAV.
The literature on rotary-wing UAV control strategies offers interesting approaches that all have specific advantages and limitations.

In \cite{Nonami2004}, Nonami \emph{et al.} disclosed an autonomous control method for an unmanned helicopter.
%In their system, the helicopter embeds a GPS for localization and other sensors to detect attitude. 
They propose a feedback control scheme based on Linear-Quadratic-Gaussian (LQG) control. 
Their solution is however specific to the control of the servo-motor of an helicopter.
Moreover this control scheme uses independent controllers for the control of longitudinal, lateral and vertical displacements which may lead to less optimal trajectories than a method processing the whole dynamics at the same time.
%Then, The X and Y controllers use a serial arrangement of separate controllers for velocity and attitude,  whereas a parallel architecture is proposed in our invention.
Unlike \cite{Nonami2004} that used an embedded GPS for localization, \cite{Troy2010} proposed a general method of closed-loop feedback control using motion capture systems allowing to track very accurately position and orientation of the controlled vehicle. However, no detailed description about the mathematical solution is given -- authors only mention LQR (linear quadratic regulation) and Kalman filtering.

%US20080097658 discloses a flight control system for an aircraft (not unmanned), which allows, among other advantages, “the aircraft to be capably guided to any waypoint with any required speed”. This system has a three control loop design (Fig. 1):
%\begin{itemize}
%\item An inner loop whose purpose is to stabilize the UAV; this control loop implements separate LQR controllers for longitudinal and lateral motion (sections [00041]-[00042], claims 1a, 2 and 5-6). A part of this system separates lateral and longitudinal control laws applying independently a LQR before regrouping (see their figure 5).
%\item An outer loop for controlling the UAV velocity, altitude, roll and yaw angles;
%\item A third steady state trim loop that specializes in maintaining the UAV as close as possible to steady state navigation (constant altitude, roll, yaw), relieving the stabilization roles of the other two loops, and therefore allowing the UAV to fly more difficult maneuvers.
%\end{itemize}
%This controller design clearly differs from the one proposed in the invention, which is architecture around two feedback control loops, one for position (not separating the longitudinal and transversal motions), and one for course control. Moreover, such a control model doesn’t address generic UAV input model as we do in this invention

In \cite{Barrientos2001}, the authors proposed a method divided in several steps. One of the steps deals with the control of the roll and pitch -- using a LQG regulation with Kalman filtering -- and another one addresses the control of the yaw.
The control strategy they propose is however limited to the hover state. Moreover, they rely on a non-generic control API, as opposed to 
%Another difference with the our ... lies in the fact that the LQG controller jointly addresses roll and pitch, whereas the UAV control in the invention is architecture around two LQG controllers for yaw and position, the yaw estimate from the first controller feeding the LQR block of the position controller at each time step.
~\cite{Krajnik2011} Krajnik \emph{et al.} that proposed a control scheme for a quadrotor drone based on the same high-level control API that we propose to use. 
The main limitation of this last approach lies in its simplicity. Using independent controllers for the control of yaw and each of the positional coordinates might, again, lead to less optimal trajectories. Furthermore, they implemented the controllers using simpler proportional or proportional-derivative controllers, as opposed to the LQG scheme that we use in this paper.
%However, the control scheme is simpler than the one proposed here, in two respects:
%First, independent controllers are designed for the control of yaw and each of the positional coordinates in the drone coordinate system which may lead to less optimal trajectories. And second, These controllers are implemented using simpler proportional or proportional-derivative controllers, as opposed to the LQG scheme that we ... in this paper.

\subsection{Camera path planning}
The problem of path planning which origin lies in the robotic field has recently evolved and impacted other fields of research such as the computer graphics community.
In the robotic field, many different approaches have been proposed. From sampling-based methods such as probabilistic roadmaps~\cite{Kavraki1994}, to force-based solutions (using potential fields\cite{Koren1991} for instance) or visual servoing~\cite{Zhang2002}, the literature offers effective solutions through a variety of different algorithm (\emph{e.g.} A*, genetic algorithm, ant algorithm, etc.). However, most of this research was restricted to ground vehicles and therefore did not fully exploit the possibilities offered by rotary-wing drones.

The literature on path planning in virtual environments was mostly inspired by the research conducted in the robotic field. However, part of this research offers new insights on the topic by addressing the specific problem of automatic cinematography. Considering that a drone is the closest embodiment of a virtual camera (due to its many degrees of freedom), this literature is essential to our research topic.
In 2008, Christie \emph{et al} \cite{Christie2008} gave a thorough analysis of the existing literature on intelligent camera control which, at the time, mostly relied on robot path planning techniques\cite{Courty2001,Li2008, Burelli2009}.
In \cite{Lino2012} however, the authors proposed an algebraic solution to the problem of placing a camera based on a set of visual constraints. This work on static camera placement was further investigated to address the challenge of camera path planning. Offline solutions, such as \cite {Lino2015} or \cite{Galvane2015} give interesting results but remain, by nature, unsuited to real-time path planning.
Closer to our research topic, \cite{Galvane2013,Galvane2014} introduced a reactive approach based on steering behaviors. Using the Prose Storyboard Language (PSL) -- a shot description language introduced in~\cite{Ronfard2013} -- as a way to define visual properties they propose a real-time solution that pushes the camera towards its optimal position while avoiding obstacles. 

\section{Servo control}
\label{servoControl}
We here propose a novel control system that uses the current measures of the dynamics to adapt the current flight control so that the resulting trajectory of the drone follows as closely as possible the computed one (see section~\ref{cinematography}). Our solution is generic and can be used with any type of rotary-wing UAV.

The modeling part described in section~\ref{translationControl} and \ref{courseControl} relies on the 2 following assumptions:
\begin{enumerate}[(i)]
\item The roll and pitch angles absolute variations are negligible %regarding the yaw angle variations. 
due to the cinematographic context that implies slow camera motion.
\item The dynamics of the drone in terms of rotation around the up axis are designed to be much slower than the dynamics of the drone in terms of translation along its forward and right axis. 
\end{enumerate}

\begin{figure}[ht]
    \centering
    \begin{subfigure}[b]{0.40\linewidth}
        \includegraphics[width=\linewidth]{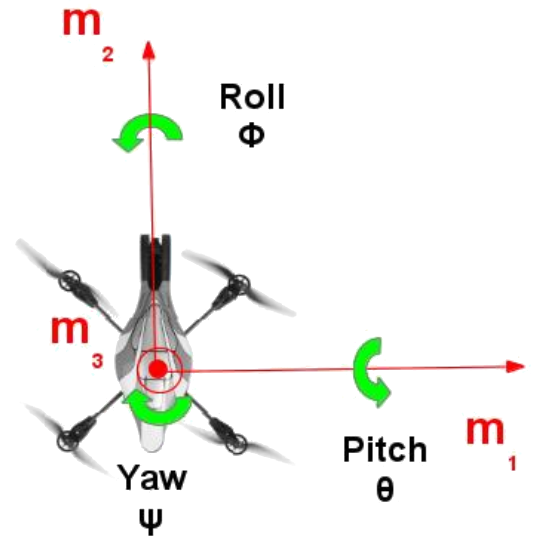}
        \caption{Drone}
    \end{subfigure}
    %\begin{subfigure}[b]{.40\linewidth} 
    %    \includegraphics[width=\linewidth]{images/mobileframe}
    %    \caption{Mobile frame}
    %\end{subfigure}
    \begin{subfigure}[b]{.40\linewidth}
        \includegraphics[width=\linewidth]{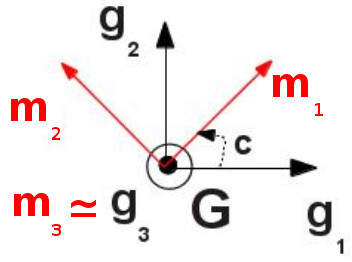}
        \caption{Global frame}
    \end{subfigure}
    %\caption{Model of a traditional rotary-wing drone (a) with the local (b) and world (c) frames }
    \caption{Model of a traditional rotary-wing drone (a) and its orientation in the world frame given by the heading $c$ (b).}
    \label{fig:model}
\end{figure}

Under assumption (i), the orientation of the drone in the global frame can be restricted to its only heading (also referred to as the course angle $c(t)$), linearly related to the yaw angle as illustrated in Figure~\ref{fig:model}. And under assumption (ii), the control of the translation and of the heading may be described by two different but coupled linear State Space Representations.

In the following, the notations introduced in Figure~\ref{fig:model} are adopted. Bold notations are also used to describe vectorial quantities whereas normal notations are used for scalar notations. Matrices are described using upper-case bold notations.

\subsection{Translation Control Model}
\label{translationControl}

Considering the previous assumptions, to model the translation control part, a continuous simple linear amortized model of the speed of the generic rotary-wing drone is initially proposed.
It can be mathematically formulated by:
\begin{equation}
\dot{v} (t) = \dot{v}_{\phi} (t) \mathbf{m_1} + \dot{v}_{\theta} (t) \mathbf{m_2} + \dot{v}_{\dot{z}} (t) \mathbf{m_3}
\end{equation}
where
\begin{equation}
\dot{v}_{\alpha}(t) = \dfrac{1}{\tau} (K_{\alpha} \alpha - v_{\alpha} (t)) \forall \alpha \in{\{ \phi, \theta, \dot{z} \}}
\end{equation}
with $K_{\alpha}$ and $\tau_{\alpha}$ being respectively the linear gain and the amortization coefficient of the model along each direction. Those gains are drone-dependent and are determined for one specific UAV making use of basic unitary tests in a calibration procedure. 

Moreover $\mathbf{m_1}$ , $\mathbf{m_2}$ and $\mathbf{m_3}$ are considered independent of the time during one step of integration (due to assumption (ii)) with
\begin{equation*}
\mathbf{m_1} =  \left\{ \begin{array}{c}
         cos(\hat{c}(t^-))\\
         sin(\hat{c}(t^-))\\
         0\end{array} \right\}% ,
\mathbf{m_2} =  \left\{ \begin{array}{c}
         -sin(\hat{c}(t^-))\\
         cos(\hat{c}(t^-))\\
         0\end{array} \right\}
%\end{equation*}
%\begin{equation*}
%and~
\mathbf{m_3} =  \left\{ \begin{array}{l}
         0\\
         0\\
         1\end{array} \right\}
\end{equation*}
where $\hat{c}(t^-)$ is an estimate of the heading of the drone just before one step of integration.
Under those conditions, it can be shown that $\dot{v} (t)$ may be written as:
\begin{equation}
\dot{\mathbf{v}}(t) = -\mathbf{T} \mathbf{v}(t) + \mathbf{M} \mathbf{T} \mathbf{K} \mathbf{u_1}(t)
\end{equation}
where 
\begin{equation*}
\mathbf{M} =  [\mathbf{m1}, \mathbf{m2}, \mathbf{m3}], 
\mathbf{T} =  \left[ \begin{array}{ccc}
         \dfrac{1}{\tau_{\phi}}& 0 & 0 \\
         0 &\dfrac{1}{\tau_{\theta}} & 0 \\
         0 & 0 & \dfrac{1}{\tau_{\dot{z}}} \end{array} \right],
\end{equation*}
\begin{equation*}
\mathbf{K} =  \left[ \begin{array}{ccc}
         K_{\phi}& 0 & 0 \\
         0 & K_{\theta} & 0 \\
         0 & 0 & K_{\dot{z}} \end{array} \right]
~and~
\mathbf{u_1} =  \left\{ \begin{array}{c}
         \phi(t)\\
         \theta(t))\\
         \dot{z}(t)\end{array} \right\}
\end{equation*}

After a discretization step, this same equation may be rewritten by:
\begin{equation}
\mathbf{v}[k+1] = (\mathbf{I_3} - \mathbf{T_D} [k]) \mathbf{v}[k] + \mathbf{M_D} [k] \mathbf{T_D} [k] \mathbf{K} \mathbf{u_1} [k]
\end{equation}
where $\mathbf{I_3}$ is the identity matrix in dimension 3, $\mathbf{M_D}[k] = 
\left\{ \begin{array}{ccc}
         cos (\hat{c}[k])&-sin (\hat{c}[k])&0\\
         sin (\hat{c}[k])&cos (\hat{c}[k])&0\\
         0&0&1\end{array} \right\}
$ and $\mathbf{T_D} [k] = \Delta[k]\mathbf{T}$ with $\Delta[k]$ being the discretisation time step at iteration k.

%We define the internal state of the State Space Representation $\mathbf{x_1} [k]$ at the iteration $k$ as the concatenation of the speed vector $\mathbf{v}[k]$ and the position vector $\mathbf{p}[k]$. At each iteration, have access to a measurement $\mathbf{y_1} [k]$ of the whole state $\mathbf{x_1} [k]$.

%By including a modeling error as an independent and identically distributed (i.i.d) additive centered Gaussian noise $\mathbf{f_1} [k]$ with known covariance matrix $\mathbf{F_1}$, and a measurement error on $\mathbf{y_1}[k]$, modeled as an i.i.d additive centered Gaussian noise $\mathbf{h_1} [k]$ of known covariance $\mathbf{H_1}$, the State Space Representation $\mathbf{x_1} [k]$ at the iteration $k+1$ is given by :

From this last equation, we derived an Explicit Discrete Time-Variant State Space Representation to model the translation control part of the UAV.
% by:
%\begin{enumerate}
%\item chosing the internal state of the State Space Representation $x_1 [k]$ at the iteration $k$ as the concatenation of the speed vector $\mathbf{v}[k]$ and the position vector $\mathbf{p}[k]$,
%\item including a modeling error as an i.i.d. additive centered Gaussian noise $\mathbf{f_1}[k]$ with known covariance matrix $\mathbf{F_1}$ and
%\item assuming than one may have access to a measurement $\mathbf{y_1}[k]$ of the whole state $\mathbf{x_1}[k]$ with a measurement error modeled as an i.i.d additive centered Gaussian noise $\mathbf{h_1[k]} of known covariance \mathbf{H_1}$,
%\end{enumerate}
%one has the following Explicit Discrete Time-Variant State Space Representation:
To do so, we first define the internal state of the State Space Representation $\mathbf{x_1} [k]$ at the iteration $k$ as the concatenation of the speed vector $\mathbf{v}[k]$ and the position vector $\mathbf{p}[k]$. We then include a modeling error as an independent and identically distributed (i.i.d) additive centered Gaussian noise $\mathbf{f_1} [k]$ with known covariance matrix $\mathbf{F_1}$,
Finally, we assume that the measurement of the whole state $\mathbf{x_1} [k]$ at each iteration $k$ is possible and given by $\mathbf{y_1} [k]$ and we model the measurement error on $\mathbf{y_1}[k]$ as an i.i.d additive centered Gaussian noise $\mathbf{h_1} [k]$ of known covariance $\mathbf{H_1}$.

The State Space Representation at the iteration $k+1$ is then given by equation~\ref{ssr} and~\ref{ssr2}:
\begin{multline}
\label{ssr}
\left\{ \begin{array}{c}
        \mathbf{v}[k+1]\\
        \mathbf{p}[k+1]\end{array} \right\} = 
\left[ \begin{array}{cc}
         \mathbf{I_3} - \mathbf{T_D} [k] & 0_3\\
         \mathbf{\Delta}\mathbf{I_3} & \mathbf{I_3}\end{array} \right]
\left\{ \begin{array}{c}
        \mathbf{v}[k]\\
        \mathbf{p}[k]\end{array} \right\}+\\
\left[ \begin{array}{c}
         \mathbf{M_D} [k]\mathbf{T_D} [k]\mathbf{K} \\
         \mathbf{0_3} \end{array} \right]
\mathbf{u_1}[k] + \mathbf{f_1}[k]
\end{multline}
\begin{equation}
\label{ssr2}
\mathbf{y_1} [k] = \mathbf{I_6} \left\{ \begin{array}{c}
        \mathbf{v}[k]\\
        \mathbf{p}[k]\end{array} \right\}
         + \mathbf{h_1} [k]
\end{equation}
Which can be generalized as:
\begin{equation}
\mathbf{x_1} [k+1] = \mathbf{A_1} [k] \mathbf{x_1} [k] + \mathbf{B_1} [k] \mathbf{u_1} [k] + \mathbf{f_1} [k]
\end{equation}
\begin{equation}
\mathbf{y_1} [k] = \mathbf{C_1} \mathbf{x_1} [k] + \mathbf{h_1} [k]
\end{equation}

%Where $\mathbf{C_1} = I_6$, $\mathbf{A_1} [k]$ accounts for the linear gain coefficients of the drone while $\mathbf{B_1} [k]$ depends on the the course of drone, the linear gain coefficients and the amortization coefficients. We fully disclosed the technical details leading to this result in the patent \cite{Fleureau2015}.

\subsection{Heading Control Model}
\label{courseControl}

%Following a similar reasoning, we define the State Space Representation $x_2 [k]$ at the iteration $k$ as the course $c[k]$.
For the heading control, we define the internal state of the  State Space Representation $x_2 [k]$ at the iteration $k$ as the course $c[k]$.
We then include a modeling error as an i.i.d additive centered Gaussian noise $f_2 [k]$ with known covariance $F_2$. And finally, assuming that the measurement of the whole state $x_2[k]$is also possible and given by $y_2[k]$, we model the measurement error as an i.i.d additive centered Gaussian noise $h_2 [k]$ of known variance $H_2$, 

The State Space Representation $x_2$ at the iteration $k+1$ is then given by : 
%The course modeling is more straightforward. By:
%\begin{enumerate}
%\item chosing the internal state of the State Space Representation $x_2[k]$ at the iteration $k$ as the course $c[k]$,
%\item including a modeling error as an i.i.d. additive centered Gaussian noise $f_2[k]$ with known variance $F_2$ and
%\item assuming than one may have access to a measurement $y_2[k]$ of the whole state $x_2[k]$ with a measurement error modeled as an i.i.d additive centered Gaussian noise $h_2[k]$ of known variance $H_2$,
%\end{enumerate}
%one has the following Explicit Discrete Time-Variant State Space Representation:
\begin{equation}
c[k+1] = 1 c[k] + K_{\dot{\psi}} \Delta[k] \dot{\psi} [k] + f_2 [k]
\end{equation}
and
\begin{equation}
y 2 [k] = 1 c[k] + h_2 [k]
\end{equation}
Which can be generalized as:
\begin{equation}
x_2 [k+1] = a_2 x_2 [k] + b_2 [k] u_2 [k] + f_2 [k]
\end{equation}
\begin{equation}
y 2 [k] = c_2 c[k] + h_2 [k]
\end{equation}

\begin{figure*}[ht]
	\centering
	\includegraphics[width=0.8\linewidth]{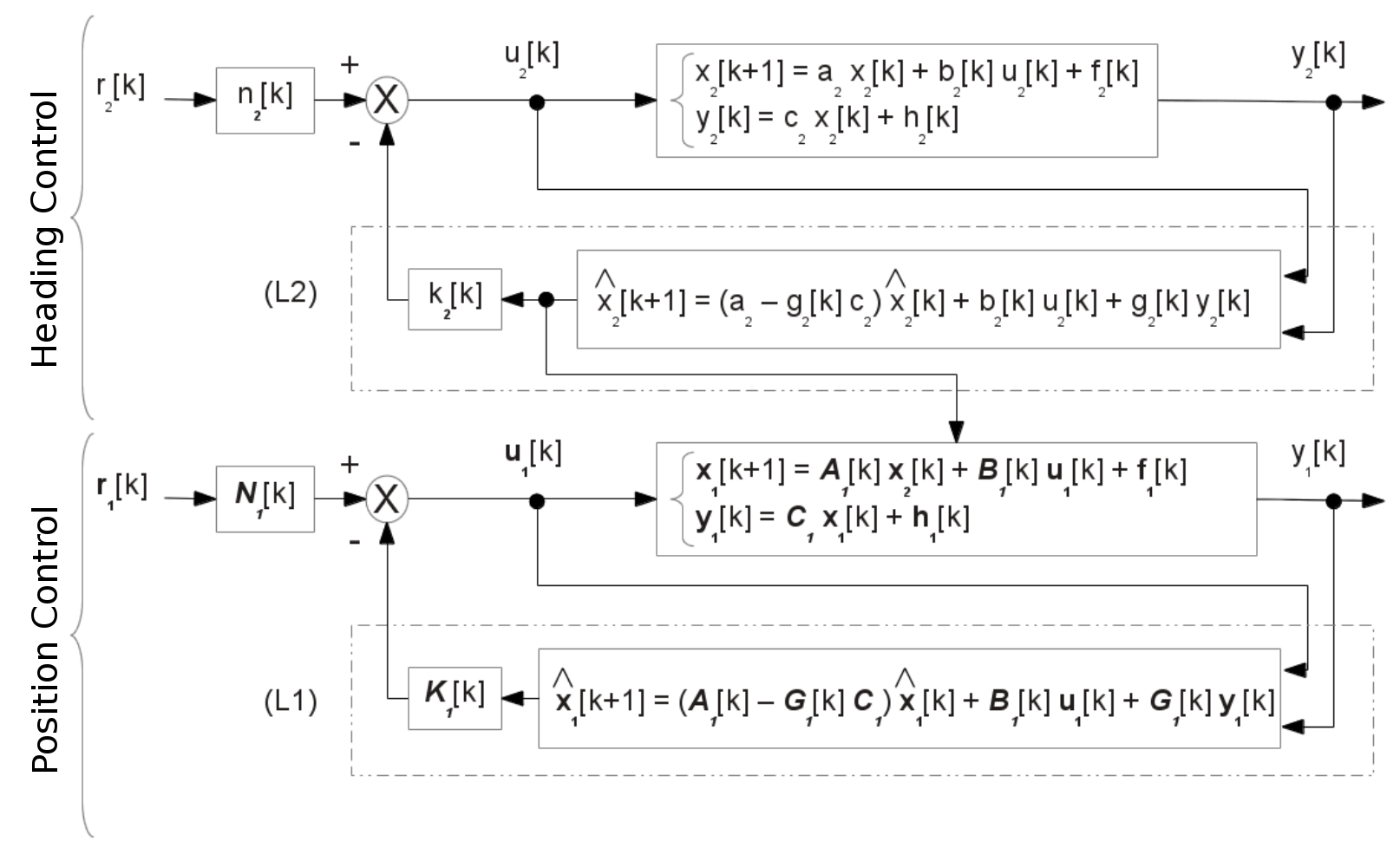}
	\caption{Control Architecture}
	\label{fig:archi}
\end{figure*}

\subsection{Control Architecture}

Given those two models, we propose a new architecture (see Figure~\ref{fig:archi}) based on two Linear Quadratic Gaussian regulators to estimate at each time step, the flight control $\mathbf{u}[k]$ so that the rotary-wing drone follows the user-defined trajectory $\mathbf{r}[k]$ with :
\begin{equation}
\mathbf{u}[k] =  \left\{ \begin{array}{ll}
         \mathbf{u_1}[k]\\
         u_2[k]\end{array} \right\} 
, \mathbf{r}[k] =  \left\{ \begin{array}{ll}
         \mathbf{r_1}[k]\\
         r_2[k]\end{array} \right\} 
         =  \left\{ \begin{array}{ll}
         \mathbf{v^r}[k]\\
         \mathbf{p^r}[k]\\
         c^r[k]\end{array} \right\} 
\end{equation}

As one can observe, the heading of the drone (top part of the block diagram in Figure~\ref{fig:archi}) is estimated independently of its position (bottom part of the block diagram) on the basis of the assumptions. 
Besides, at each time step, the matrix $\mathbf{B_1} [k]$ is built making use of the heading $\hat{x}_2 [k+1]$ estimated from the kalman filter.
The command $\mathbf{u}[k]$ at each time step $k$ is computed from:
\begin{equation}
\mathbf{u}[k] =  \left\{ \begin{array}{ll}
         \mathbf{u_1}[k]\\
         u_2[k]\end{array} \right\} 
         =  \left\{ \begin{array}{ll}
         \mathbf{N} [k] \mathbf{r_1} [k] - \mathbf{K_1} [k] \mathbf{\hat{x}_1} [k]\\
         n_2 [k] r_2 [k] - k_2 [k] \hat{x}_2 [k]\end{array} \right\} 
\end{equation}
where ($\mathbf{N_1} [k]$, $\mathbf{K_1} [k]$) and ($n_2 [k]$, $k_2 [k]$) are the pre-filter and Full State Feedback gain of the position control block and of the heading control block respectively, and $\mathbf{\hat{x}_1} [k]$ and $\hat{x}_2 [k]$ are the estimates of the states $\mathbf{x_1} [k]$ and $x_2 [k]$ respectively, computed from the associated kalman filters.
%Further information on $\mathbf{N_1} [k]$, $n_2 [k]$, $\mathbf{\hat{x}_1} [k]$ and $\hat{x}_2 [k]$ are etailed in the patent \cite{Fleureau2015}.

Following a LQG strategy, the gain $\mathbf{K_1}[k]$ is computed as the solution of a Discrete LQG regulator with infinite horizon, \emph{i.e.}, $\mathbf{K_1} [k] = (\mathbf{R} + \mathbf{B_1} [k]^T \mathbf{P}[k] \mathbf{B_1} [k])^{-1} \mathbf{B_1} [k]^T \mathbf{P}[k] \mathbf{A_1} [k]$ \\where $\mathbf{P}[k]$ is the solution of the Discrete Algebraic Ricatti Equation $\mathbf{P}[k] = \mathbf{Q} + \mathbf{A_1} [k] (\mathbf{P}[k] - \mathbf{P}[k] \mathbf{B_1} [k] (\mathbf{R} + \mathbf{B_1} [k]^T \mathbf{P}[k] \mathbf{B_1} [k])^{-1} \mathbf{B_1} [k]^T \mathbf{P}[k]) \mathbf{A_1} [k]$. Such a solution may be computed by the means of specific algorithm based on a Schur decomposition for instance. The matrices Q and R are two user-defined weighting matrices which allow to control the quality of the regulation. They are initially set to the identity matrix. Then, ``increasing'' $\mathbf{R}$ (multiplying $\mathbf{R}$ by a positive scalar) allows to smooth the dynamics of the control flight $(u_1)$, whereas ``increasing'' $\mathbf{Q}$ allows to smooth the flight dynamics ($x_1$).
The pre-filter $\mathbf{N_1} [k]$ is chosen so that the Full State Feedback converge toward $r_1 [k]$. One can especially show that a choice of $\mathbf{N_1} [k] = (\mathbf{B_1} [k]^T \mathbf{B_1} [k])^{-1} \mathbf{B_1} [k]^T ( (\mathbf{I_6} + \mathbf{B_1} [k] \mathbf{K_1} [k]) - \mathbf{A_1}[k]) $ is appropriate.

$\hat{x}_1 [k]$ is the estimate of $x_1 [k]$ from a standard Discrete Non-Stationnary Kalman Filter with a gain $\mathbf{G_1} [k]$ at the step $k$ which update is done by the means of the covariance matrices $\mathbf{F_1}$ and $\mathbf{H_1}$, and of the translation control model.

The gain $k_2[k]$ is computed so that the dynamics of the state $x_2[k]$ are slow enough to be compatible with assumption (i). To have an attenuation $\gamma$ of the control error after a time $\tau$, one could especially chose $k_2[k] = exp\left(\dfrac{\overline{\Delta[k]} log (\gamma) }{\tau-\overline{\Delta[k]}}\right)$ where $\overline{\Delta[k]}$ is the mean value of the time step of the process.
The pre-filter $n_2[k]$ is chosen so that the Full State Feedback converge toward $r_2 [k]$. One can show that the choice $n_2 [k] = k_2 [k]$ is appropriate.

$\hat{x}_2 [k]$ is the estimate of $x_2 [k]$ from a standard Discrete Non-Stationnary Kalman Filter with a gain $g_1 [k]$ at the step k which update is done making use of the variances $f_1$ and $h_1$, and of the heading control model.

\section{Drone and cinematography}
\label{cinematography}

%After detailing the process dealing with the servo-control of the drone, we now go through the trajectory computation process. 
In this section we address the problem of automatically computing camera trajectories.
The goal is to provide the proper data to the controller to ensure that the drone follows a cinematographic trajectory. 
As seen in section~\ref{servoControl} the controller takes as input three components: the position, the orientation and the speed. 
This information can either be manually generated -- using third party softwares such as 3D modelers -- or can be automatically computed from high level specifications. In this section we address the automatic approach and propose a three-stages process.

The first step consists of interacting with a user and extract camera configurations from high-level specifications. The camera configurations are defined relatively to the actors and expressed through a 2D-parametric representation, using two types of manifold surfaces: a spherical surface (for one actor) or a toric-shaped surface (for two actors ; see~\cite{Lino2012}). Each of the parameters defining the manifold surfaces are associated with visual properties (\emph{e.g.} shot size, position on the screen, profile angle, etc.).
To provide high-level control of these properties we use the PSL \cite{Ronfard2013} (examples of PSL commands are given in Figures~\ref{fig:scenario1} and ~\ref{fig:scenario2}) and map each of the keywords of its grammar to specific values of the parametric representation. For instance, in a one-actor situation, the shot size is associated with the radius of the sphere while the profile angle of an actor gives the position of the camera on the sphere.% and the position on the screen defines its orientation.

\begin{figure}[ht]
    \centering
    \begin{subfigure}[b]{.3\linewidth}
        \includegraphics[width=\linewidth]{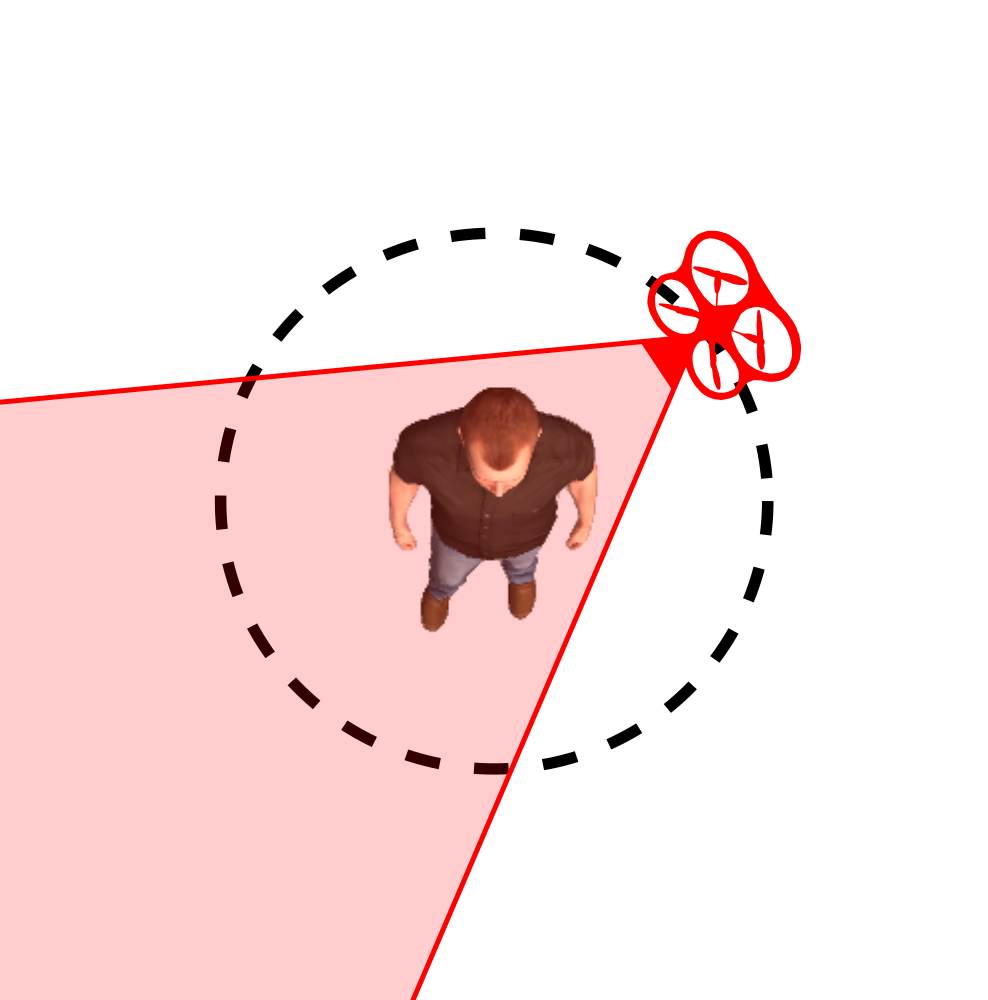}
        \caption{}
    \end{subfigure}
    \begin{subfigure}[b]{0.3\linewidth}
        \includegraphics[width=\linewidth]{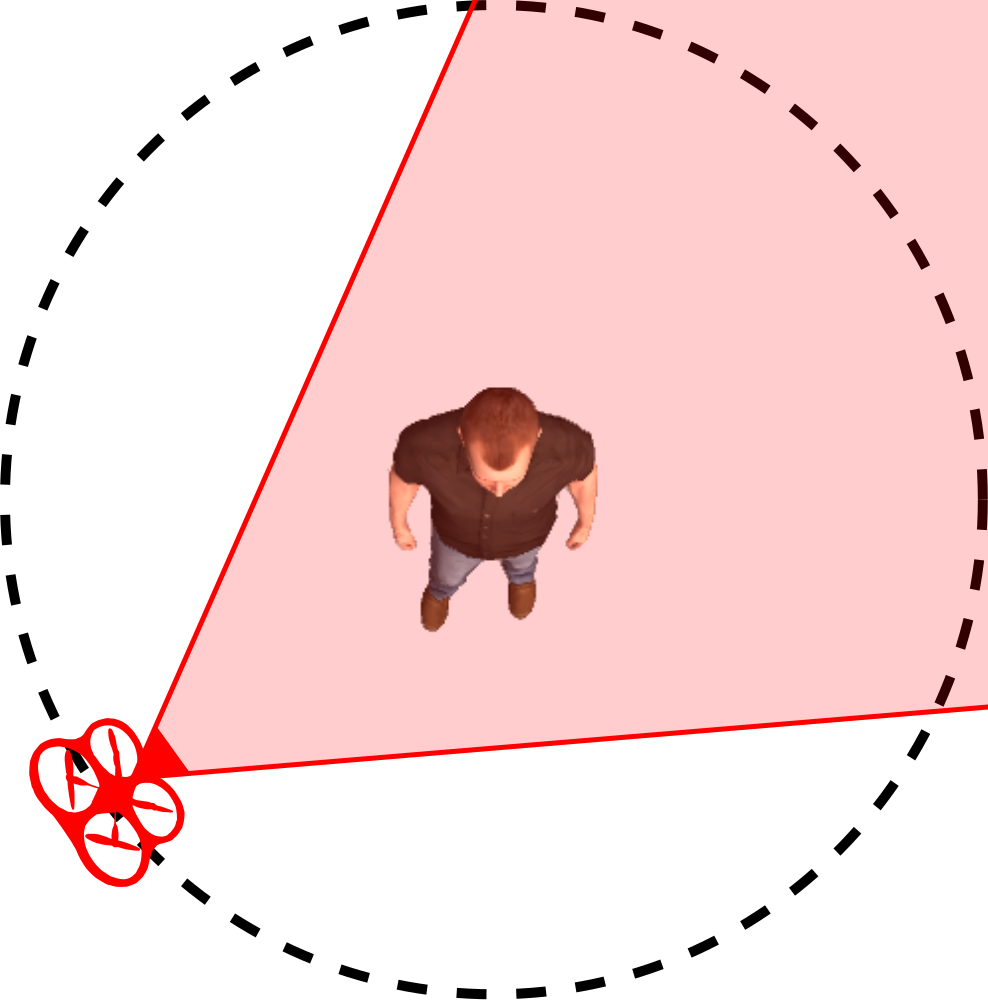}
        \caption{}
    \end{subfigure}
    \begin{subfigure}[b]{.3\linewidth}
        \includegraphics[width=\linewidth]{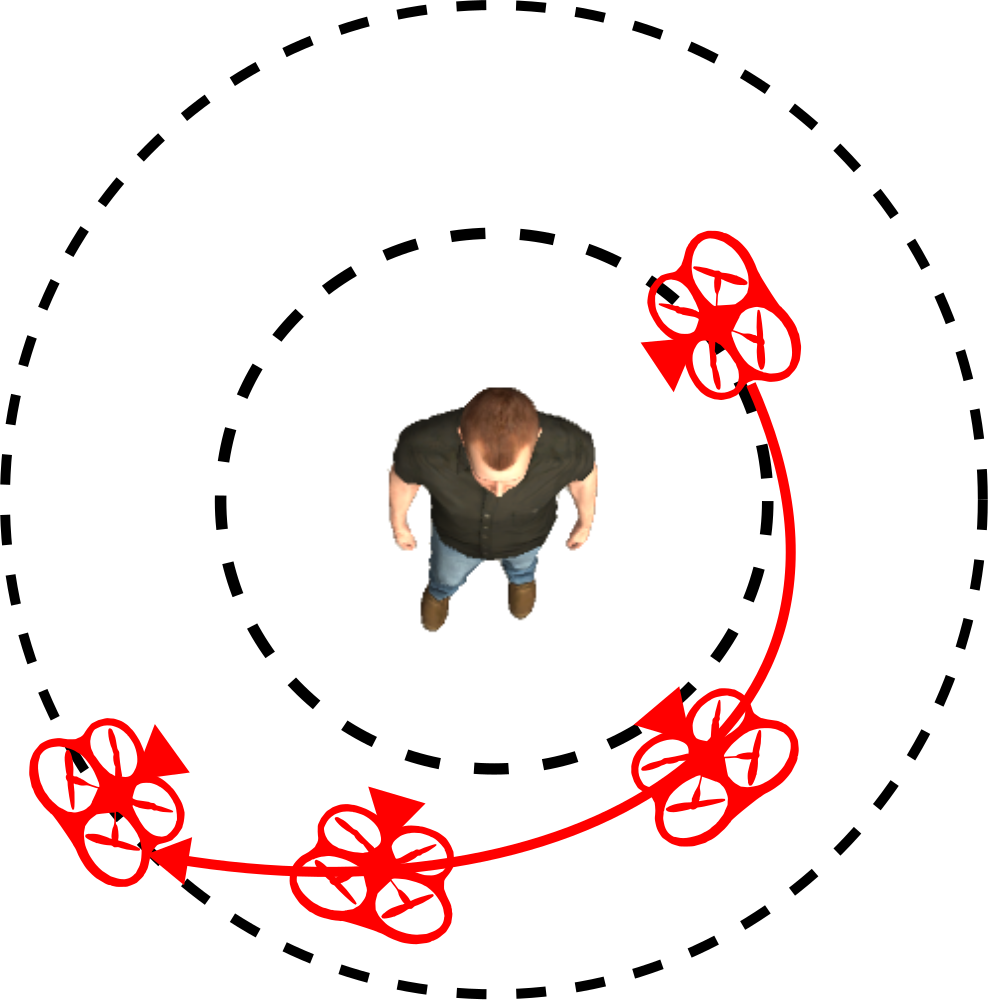}
        \caption{}
    \end{subfigure}
    \caption{Path generated from an initial (a) to a final camera configuration (b) by interpolating on the manifold space (c).}
    \label{fig:transition}
\end{figure}

The second step of the process lies in the creation of a path, given an initial camera placement and a final camera configuration (provided by the user). The path is computed by interpolating the parameters in the manifold space to provide smooth motion. Figure~\ref{fig:transition} illustrates a transition between two camera specifications. Each of the visual properties are linearly interpolated to compose a smooth and cinematographic path.

Finally, given the path to be followed by the camera, the last step consists of computing the position and speed of the camera at each time step. The path being defined as an interpolation between two camera configurations, given the starting time of the shot and its duration, it is straightforward to compute the optimal position at each time iteration and deduce the speed. Such solution however does not account for the aesthetic of the shot and would produce abrupt camera motion (with strong acceleration and irregular speed). To solve the problem, we use the steering behaviors introduced in~\cite{Galvane2013} to gradually \emph{push} the camera towards its optimal position and generate a smooth motion along the path. This solution rely on the acceleration, speed and position of the simulated drone and therefore provide all the information needed by the controller.

\section{Results}
\label{results}
To test our approach, we conducted several shooting experiments in an indoor environment.
These first results were obtained with a \emph{Parrot AR.Drone}, one of the most popular quadrotor on the market. 
In addition to its low cost this drone is equiped with a static camera and provide the generic flight controls described in section~\ref{servoControl}.
The measures of the flight dynamics were computed with a full \emph{OptiTrack} setup.
All the results given in this section are further illustrated in the companion video\footnote{https://vimeo.com/157994891}.

\begin{figure}[ht]
    \centering
    %\begin{subfigure}[b]{0.49\linewidth}
    %    \includegraphics[width=\linewidth]{images/sq1}
        %\caption{}
	%\vspace*{-3mm}
    %\end{subfigure}
    \begin{subfigure}[b]{.32\linewidth}
        \includegraphics[width=\linewidth]{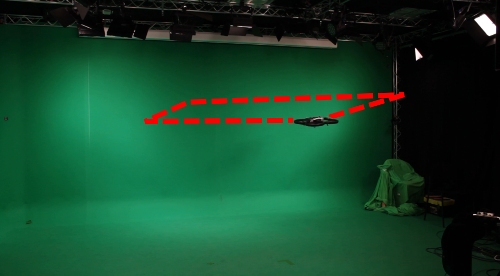}
        \caption{}
    \end{subfigure}
    \begin{subfigure}[b]{.32\linewidth}
        \includegraphics[width=\linewidth]{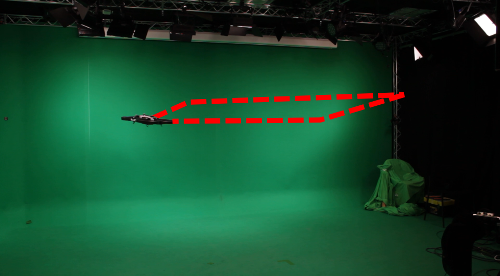}
        \caption{}
    \end{subfigure}
    \begin{subfigure}[b]{0.32\linewidth}
        \includegraphics[width=\linewidth]{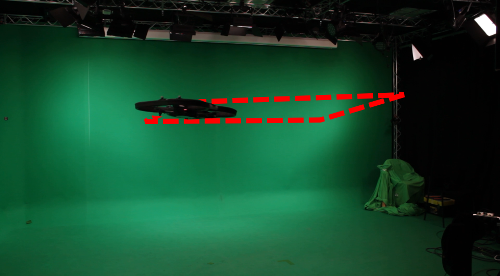}
        \caption{}
    \end{subfigure}
    
    \begin{subfigure}[b]{0.32\linewidth}
        \includegraphics[width=\linewidth]{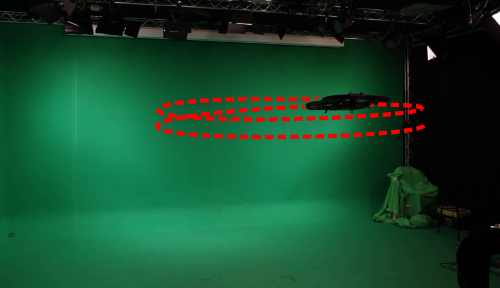}
        \caption{}
    \end{subfigure}
    \begin{subfigure}[b]{.32\linewidth}
        \includegraphics[width=\linewidth]{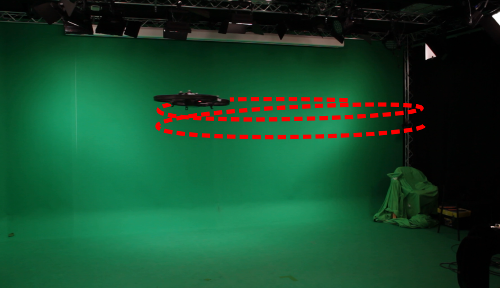}
        \caption{}
    \end{subfigure}
    \begin{subfigure}[b]{.32\linewidth}
        \includegraphics[width=\linewidth]{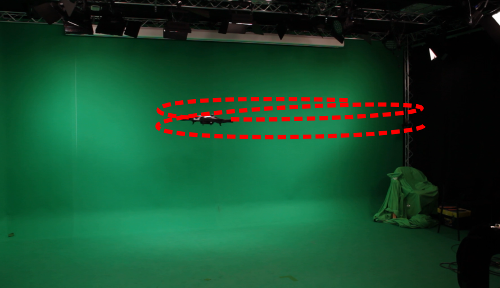}
        \caption{}
   \end{subfigure}
    \caption{The drone executes predefined trajectories such as a square (a,b,c) or a helix (d,e,f).}
    \label{fig:predefined}
\end{figure}

The first set of experiments aimed at validating the servo-control mechanism by executing manually crafted trajectories. Figure~\ref{fig:predefined} illustrates the behavior of the drone while performing a square and a helix trajectory. The drone manages to closely follow the path and keep up with the specified timing of the shot. 
This capacity to reproduce predefined trajectories is essential for shooting purposes. 
It gives directors the possibility to design complex camera trajectories beforehand and let the drone autonomously take care of the shooting.
The helix trajectory for instance is a classical shot often used in movies.% to reveal a new character for instance. %to turn around an actor while ...

%\begin{figure}[ht]
%    \centering
%    \begin{subfigure}[b]{0.49\linewidth}
%        \includegraphics[width=\linewidth]{images/h1}
%        %\caption{}
%\vspace*{-3mm}
%    \end{subfigure}
%    \begin{subfigure}[b]{.49\linewidth}
%        \includegraphics[width=\linewidth]{images/h2}
%        %\caption{}
%\vspace*{-3mm}
%    \end{subfigure}
%    \begin{subfigure}[b]{.49\linewidth}
%        \includegraphics[width=\linewidth]{images/h3}
%        %\caption{}
%   \end{subfigure}
%    \begin{subfigure}[b]{0.49\linewidth}
%        \includegraphics[width=\linewidth]{images/h4}
%        %\caption{}
%    \end{subfigure}
%    \caption{The drone executes a predefined helix trajectory.}
%    \label{fig:helix}
%\end{figure}

\begin{figure}[ht]
    \centering
    \begin{subfigure}[b]{0.32\linewidth}
        \includegraphics[width=\linewidth]{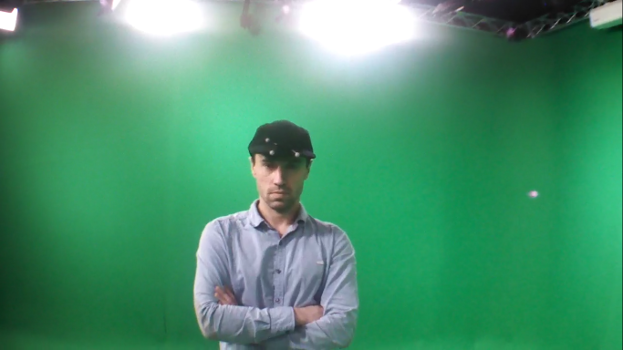}
        \caption{}
    \end{subfigure}
    \begin{subfigure}[b]{.32\linewidth}
        \includegraphics[width=\linewidth]{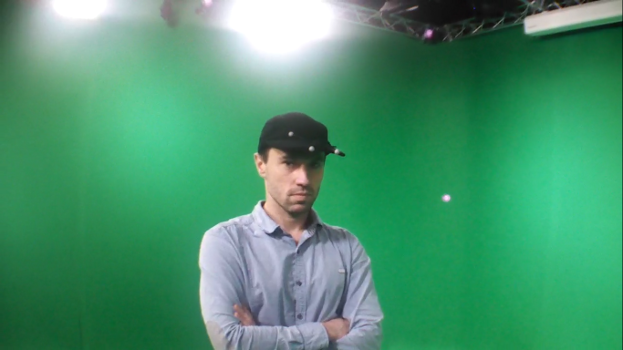}
        \caption{}
    \end{subfigure}
    \begin{subfigure}[b]{.32\linewidth}
        \includegraphics[width=\linewidth]{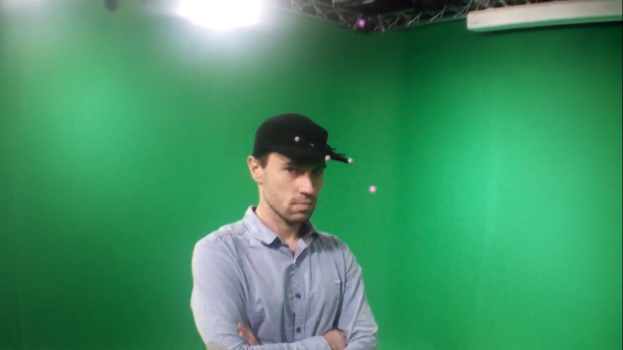}
        \caption{}
    \end{subfigure}

    \begin{subfigure}[b]{0.32\linewidth}
        \includegraphics[width=\linewidth]{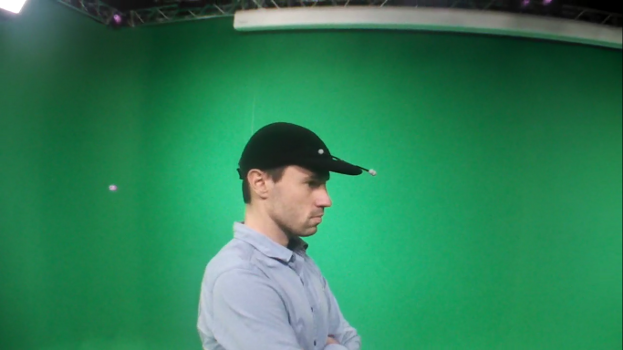}
        \caption{}
    \end{subfigure}
    \begin{subfigure}[b]{.32\linewidth}
        \includegraphics[width=\linewidth]{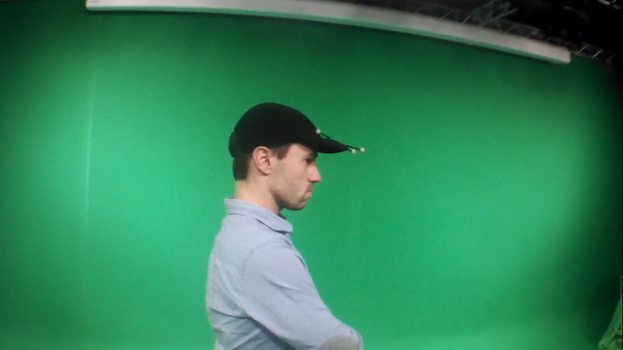}
        \caption{}
    \end{subfigure}
    \begin{subfigure}[b]{.32\linewidth}
        \includegraphics[width=\linewidth]{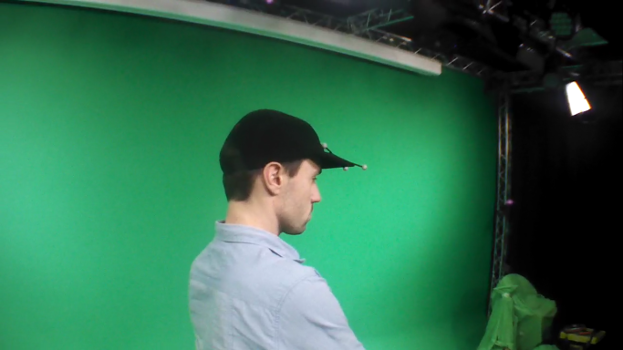}
        \caption{}
    \end{subfigure}
    \caption{The drone transitions from an initial configuration \emph{``MS on A front''} (a) towards a final PSL specification \emph{``MS on A 34backright''} (f)}
    \label{fig:scenario1}
\end{figure}

The next experiments intended to fully test our autonomous system for the shooting of cinema scenes with one or two moving actors.
In Figure~\ref{fig:scenario1}, we show an example of shot transition between two camera configurations. The user provided a high-level description of the shot in the form of a PSL sentence and our system autonomously generated an aesthetic trajectory for the drone to transition from its current position to the desired configuration.

\begin{figure}[ht]
    \centering
    \begin{subfigure}[b]{0.32\linewidth}
        \includegraphics[width=\linewidth]{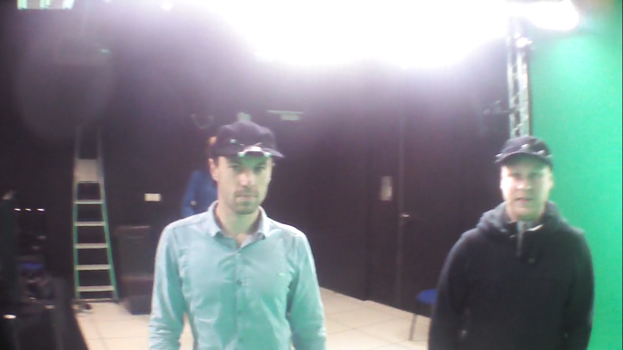}
        \caption{}
    \end{subfigure}
    \begin{subfigure}[b]{.32\linewidth}
        \includegraphics[width=\linewidth]{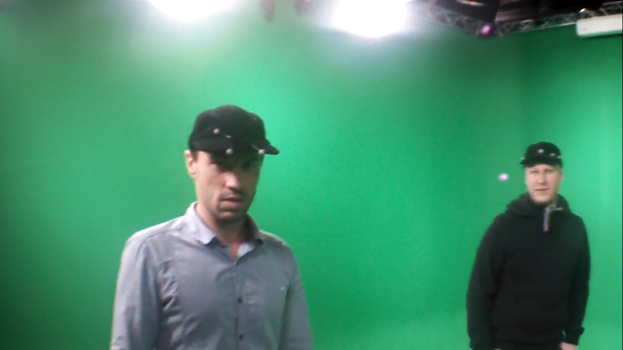}
        \caption{}
    \end{subfigure}
    \begin{subfigure}[b]{.32\linewidth}
        \includegraphics[width=\linewidth]{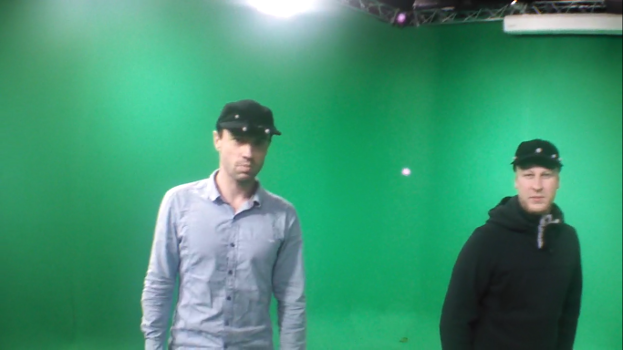}
        \caption{}
    \end{subfigure}

    \begin{subfigure}[b]{0.32\linewidth}
        \includegraphics[width=\linewidth]{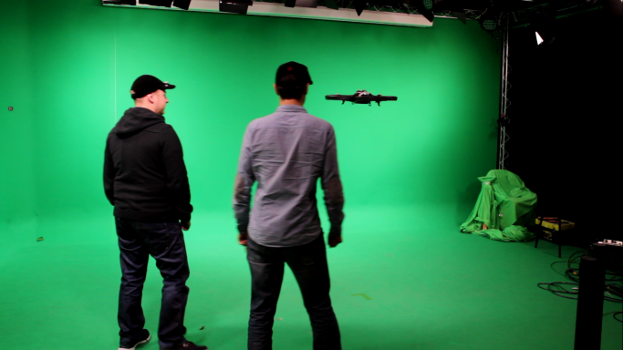}
        \caption{}
    \end{subfigure}
    \begin{subfigure}[b]{.32\linewidth}
        \includegraphics[width=\linewidth]{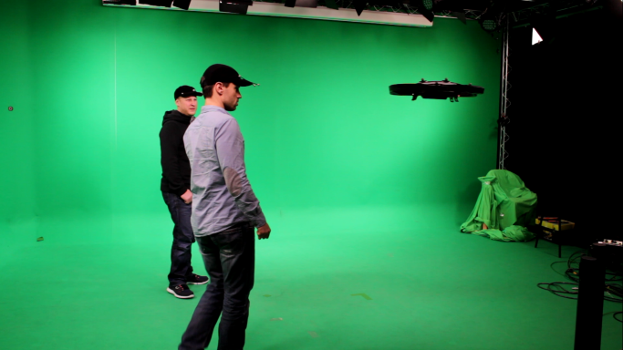}
        \caption{}
    \end{subfigure}
    \begin{subfigure}[b]{.32\linewidth}
        \includegraphics[width=\linewidth]{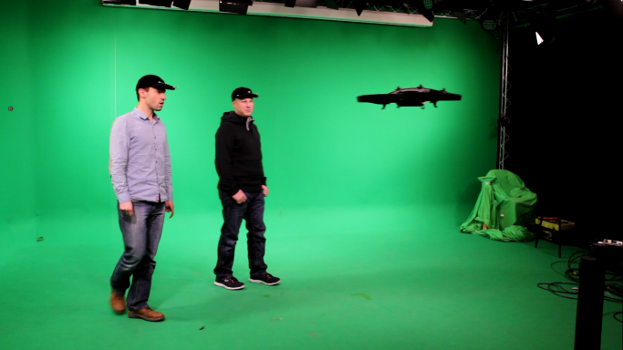}
        \caption{}
    \end{subfigure}
    \caption{The drone autonomously maintains a given framing over two moving actors. (a,b,c) show the resulting shot and (d,e,f) give an overview of the scene to show the movement of the actors and the drone.}
    \label{fig:scenario2}
\end{figure}

As illustrated in Figure~\ref{fig:scenario2}, the drone can also handle multiple actors. It is able to maintain the framing (\emph{i.e.} the screen composition) or transition between camera specifications over two moving actors.
Other scenarios were also tested and the results can be found in the joined video.

\section{Limitation and Future work}
\label{limitations}

Currently, our system only manages a single drone. Handling multi-camera setup using UAV is a challenging endeavor with tremendous potential that clearly deserves further investigation in terms of camera path planning.
Testing the system with other setup is also a priority. Eventhough the \emph{Parrot AR.Drone} offers many advantages, it suffers from many limitations -- such as poor stability, low battery autonomy and lack of mobility of the camera (that make high angle camera shots impossible).
Taking advantage of our generic servo-control mechanism, part of our future work will be dedicated to experimenting our system with other rotary-wing drones offering more possibilities (with camera gimbals for instance). 

\section{Conclusion}
%Using ... our solution is, by nature, drone-independent and could be used with any rotary-wing UAV that offer generic flight control ...

In this paper, we addressed a problem never explored before -- the autonomous capture of cinema scenes in real-time using drones -- and proposed a generic solution to this complex endeavor.
After proposing a compound and coupled model that gives a first-order temporal relation between the flight control of the drone and its dynamics, along with the global control architecture integrating the two distinct State Space Representation, we proposed a path planning approach dedicated to the live capture of cinema scenes.

This research constitute a first step towards the creation of autonomous tools for the cinema industry. Such tools would most certainly greatly benefits movie-makers, ranging from amateurs and their low-budget productions to professionals for pre-production purposes.
%It especially includes i) an Explicit Discrete Time-Variant State Space Representation of the translation control of the UAV and ii) an Explicit Discrete Time-Variant State Space Representation of the course control of the UAV.

%The global control architecture integrating the two distinct Explicit Discrete Time-Variant State Space Representation -- one for the translation and one for the course -- and a Full State Feedback strategy gives a generic solution that could be used with any rotary-wing drone offering generic flight-control.

%we proposed a complete platform for the autonomous shooting of cinema scenes using drones.

%ACKNOWLEDGMENTS are optional
%\section{Acknowledgments}

\bibliographystyle{abbrv}
\bibliography{Dronet2016}  % sigproc.bib is the name of the Bibliography in this case

\begin{thebibliography}{10}

\bibitem{Barrientos2001}
A.~Barrientos, I.~Aguirre, J.~Del-Cerro, and P.~Portero.
\newblock Lqg vs pid for attitude control of a for unmanned aerial vehicle in
  hover.
\newblock In {\em Proceedings of the 10th International Conference on Advanced
  Robotics 2001}, ICAR 2001, pages 599--604, Budapest, Hungary, 2001.

\bibitem{Burelli2009}
P.~Burelli and A.~Jhala.
\newblock Dynamic artificial potential fields for autonomous camera control.
\newblock In {\em Artificial Intelligence In Interactive Digitale Entertainment
  Conference}, Palo Alto, 2009. AAAI.

\bibitem{Christie2008}
M.~Christie, P.~Olivier, and J.-M. Normand.
\newblock Camera control in computer graphics.
\newblock {\em Computer Graphics Forum}, 27(8):2197--2218, 2008.

\bibitem{Courty2001}
N.~Courty and E.~Marchand.
\newblock Computer animation: a new application for image-based visual
  servoing.
\newblock In {\em Robotics and Automation, 2001. Proceedings 2001 ICRA. IEEE
  International Conference on}, volume~1, pages 223--228 vol.1, 2001.

\bibitem{Galvane2015}
Q.~Galvane, M.~Christie, C.~Lino, and R.~Ronfard.
\newblock {Camera-on-rails: Automated Computation of Constrained Camera Paths}.
\newblock In {\em { ACM SIGGRAPH Conference on Motion in Games}}, Paris,
  France, 2015.

\bibitem{Galvane2013}
Q.~Galvane, M.~Christie, R.~Ronfard, C.-K. Lim, and M.-P. Cani.
\newblock {Steering Behaviors for Autonomous Cameras}.
\newblock In {\em {MIG 2013 - ACM SIGGRAPH conference on Motion in Games}},
  pages 93--102, Dublin, Ireland, Nov. 2013. {ACM}.

\bibitem{Galvane2014}
Q.~Galvane, R.~Ronfard, M.~Christie, and N.~Szilas.
\newblock {Narrative-Driven Camera Control for Cinematic Replay of Computer
  Games}.
\newblock In {\em {Motion In Games}}, Los Angeles, United States, Nov. 2014.

\bibitem{Kavraki1994}
L.~Kavraki, P.~Svestka, J.~Latombe, and M.~Overmars.
\newblock Probabilistic roadmaps for path planning in high-dimensional
  configuration spaces.
\newblock Technical report, Stanford, CA, USA, 1994.

\bibitem{Koren1991}
Y.~Koren and J.~Borenstein.
\newblock Potential field methods and their inherent limitations for mobile
  robot navigation.
\newblock In {\em In Proceedings of the IEEE International Conference on
  Robotics and Automation}, pages 1398--1404, 1991.

\bibitem{Krajnik2011}
T.~Krajn{\'\i}k, V.~Von{\'a}sek, D.~Fi{\v{s}}er, and J.~Faigl.
\newblock Ar-drone as a platform for robotic research and education.
\newblock In {\em Research and Education in Robotics-EUROBOT 2011}, pages
  172--186. Springer Berlin Heidelberg, 2011.

\bibitem{Li2008}
T.-Y. Li and C.-C. Cheng.
\newblock {Real-Time Camera Planning for Navigation in Virtual Environments}.
\newblock In {\em SG '08: Proceedings of the 9th international symposium on
  Smart Graphics}, pages 118--129, Berlin, Heidelberg, 2008. Springer-Verlag.

\bibitem{Lino2012}
C.~Lino and M.~Christie.
\newblock Efficient composition for virtual camera control.
\newblock In {\em Proceedings of the Eurographics Symposium on Computer
  Animation}, pages 65--70. Eurographics Association, 2012.

\bibitem{Lino2015}
C.~Lino and M.~Christie.
\newblock Intuitive and efficient camera control with the toric space.
\newblock {\em ACM Transactions on Graphics (TOG)}, 34(4):82, 2015.

\bibitem{Nonami2004}
K.~Nonami, J.~Shin, D.~Fujiwara, K.~Hazawa, and K.~Matsusaka.
\newblock Autonomous control method for a small, unmanned helicopter, Dec.~9
  2004.
\newblock US Patent App. 10/786,051.

\bibitem{Ronfard2013}
R.~Ronfard, V.~Gandhi, and L.~Boiron.
\newblock {The Prose Storyboard Language: A Tool for Annotating and Directing
  Movies}.
\newblock In {\em {2nd Workshop on Intelligent Cinematography and Editing part
  of Foundations of Digital Games - FDG 2013}}, Chania, Crete, Greece.

\bibitem{Troy2010}
J.~J. Troy, C.~A. Erignac, and P.~Murray.
\newblock Closed-loop feedback control using motion capture systems, 2010.

\bibitem{Zhang2002}
H.~Zhang and J.~Ostrowski.
\newblock Visual motion planning for mobile robots.
\newblock {\em Robotics and Automation, IEEE Transactions on}, 18(2):199--208,
  Apr 2002.

\end{thebibliography}
% ACM needs 'a single self-contained file'!
%\balancecolumns
%\appendix
%Appendix A

%\balancecolumns % GM June 2007
% That's all folks!
\end{document}